\pdfoutput=1

\documentclass[11pt]{article}

\usepackage[final]{acl}

\usepackage{times}
\usepackage{latexsym}

\usepackage[T1]{fontenc}

\usepackage[utf8]{inputenc}

\usepackage{microtype}

\usepackage{inconsolata}

\usepackage{graphicx}

\usepackage{blindtext}
\usepackage{booktabs}
\usepackage{latexsym}
\usepackage{enumerate}
\usepackage{multirow}
\usepackage{multicol}
\usepackage{booktabs}
\usepackage{tabularray}
\usepackage{comment}
\usepackage{amsmath}
\usepackage{adjustbox}
\usepackage{subcaption}
\usepackage[most]{tcolorbox}
\usepackage{xcolor,colortbl}
\usepackage{algorithm}
\usepackage{algorithmic}
\usepackage{tikz}
\usetikzlibrary{arrows.meta,positioning,calc,shapes.misc,shapes.geometric,shapes.symbols}
\usetikzlibrary{arrows.meta,positioning,calc,shapes.misc,shapes.symbols}
\usetikzlibrary{shapes.symbols}

\usepackage{soul}
\usepackage{threeparttable}

\definecolor{lightbrown}{RGB}{243, 236, 226}
\definecolor{userA}{RGB}{213,232,212}   
\definecolor{userB}{RGB}{255,242,204}   
\definecolor{userC}{RGB}{207,226,243}   
\definecolor{userD}{RGB}{244,204,204}   
\definecolor{userE}{RGB}{234,234,234}   
\definecolor{threadArrow}{RGB}{70,110,190} 
\definecolor{myBlue}{RGB}{84,124,204}
\definecolor{myLightBlue}{RGB}{197,214,242}
\definecolor{myPeach}{RGB}{244,214,198}
\definecolor{myGreen}{RGB}{202,224,195}
\definecolor{myCloud}{RGB}{252,244,205}
\definecolor{myDash}{RGB}{42,73,140}

\title{ThreadSumm: Summarization of Nested Discourse Threads Using Tree of Thoughts}

\author{Olubusayo Olabisi \\
  PortNLP Lab, Portland State University\\
  \texttt{oolabisi@pdx.edu} \\\And
  Ekata Mitra \\
  Portland State University\\
  \texttt{ekata@pdx.edu} \\\And
  Ameeta Agrawal \\
  Portland State University\\
  \texttt{ameeta@pdx.edu} \\}

\author{Olubusayo Olabisi, Ekata Mitra, Ameeta Agrawal \\
         Department of Computer Science, Portland State University, USA\\
         \texttt{\{oolabisi, ekata, ameeta\}}@pdx.edu}

\begin{document}
\maketitle
\begin{abstract}

Summarizing deeply nested discussion threads requires handling interleaved replies, quotes, and overlapping topics, which standard LLM summarizers struggle to capture reliably. We introduce \textit{ThreadSumm}, a multi-stage LLM framework that treats thread summarization as a hierarchical reasoning problem over explicit aspect and content unit representations. Our method first performs content planning via LLM-based extraction of discourse aspects and Atomic Content Units, then applies sentence ordering to construct thread-aware sequences that surface multiple viewpoints rather than a single linear strand. On top of these interpretable units, \textit{ThreadSumm} employs a Tree of Thoughts search that generates and scores multiple paragraph candidates, jointly optimizing coherence and coverage within a unified search space. With this multi-proposal and iterative refinement design, we show improved performance in generating logically structured summaries compared to existing baselines, while achieving higher aspect retention and opinion coverage in nested discussions.

\end{abstract}

\section{Introduction}

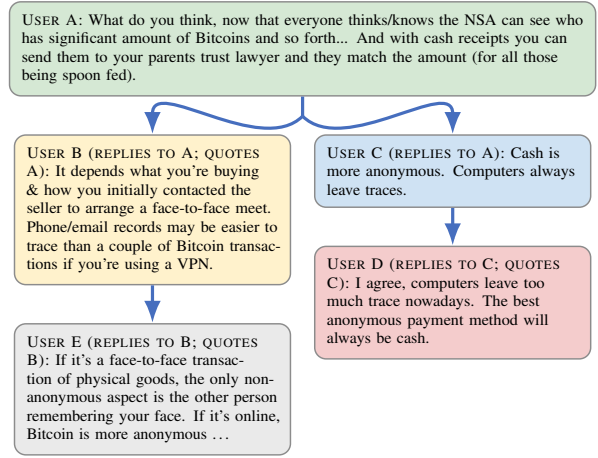
\begin{figure}
\centering
\begin{tikzpicture}[
  font=\tiny,
  box/.style={
    draw=black!45,
    rounded corners=1.5mm,
    line width=0.55pt,
    inner sep=1.6mm,
    align=left
  },
  arr/.style={line width=1.2pt, color=threadArrow}
]

\node[box, fill=userA, text width=.96\columnwidth] (A) {%
\textsc{User A:} What do you think, now that everyone thinks/knows the NSA can see who
has significant amount of Bitcoins and so forth...  And with cash receipts you can
send them to your parents trust lawyer and they match the amount (for all those
being spoon fed).
};

\node[box, fill=userB, text width=.43\columnwidth,
      below left=5mm and 1.5mm of A.south] (B) {%
\textsc{User B (replies to A; quotes A):} It depends what you're buying \& how you
initially contacted the seller to arrange a face-to-face meet. Phone/email
records may be easier to trace than a couple of Bitcoin transactions if you're
using a VPN.
};

\node[box, fill=userC, text width=.43\columnwidth,
      below right=5mm and 1.5mm of A.south] (C) {%
\textsc{User C (replies to A):} Cash is more anonymous. Computers always leave
traces.
};

\node[box, fill=userE, text width=.43\columnwidth, below=5mm of B] (E) {%
\textsc{User E (replies to B; quotes B):} If it's a face-to-face transaction of
physical goods, the only non-anonymous aspect is the other person remembering
your face. If it's online, Bitcoin is more anonymous \ldots
};

\node[box, fill=userD, text width=.43\columnwidth, below=5mm of C] (D) {%
\textsc{User D (replies to C; quotes C):} I agree, computers leave too much trace
nowadays. The best anonymous payment method will always be cash.
};

\coordinate (split) at ([yshift=-2.6mm]A.south);

\draw[arr] (A.south) -- (split);

\draw[arr,-Latex] (split) to[out=225,in=90] (B.north);  
\draw[arr,-Latex] (split) to[out=315,in=90] (C.north);  

\draw[arr,-Latex] (B.south) -- (E.north);               
\draw[arr,-Latex] (C.south) -- (D.north);               

\end{tikzpicture}
\caption{Nested thread structure illustrating reply and quote relations (A$\rightarrow$B$\rightarrow$E and A$\rightarrow$C$\rightarrow$D).}
\label{fig:thread-example}
\end{figure}

Discussion forums are known to have nested structures, as users respond to one another, quoting or replying to topics of interest, {leading to a long thread of content associated with a single post. These multi-turn conversations on discussion forums have additional layers of complexity required to navigate the discussion threads \cite{liu2025needling}. There is a tendency for threads to attract off-topic replies that get interleaved with the author’s own continuation posts, burying the main content. The nested discourse threads can be represented as graphs connected by reply, retweet, and quote edges, making it inherently harder to follow linearly compared to traditional structured documents. The fragmented and tree-like structure is further branched out by mixed reply types such as reposts, quoted posts and replies} as shown in Figure \ref{fig:thread-example}, and navigating these branches to follow one coherent strand {is crucial for summarization}.

Traditional approaches to abstractive text summarization largely focus on optimizing the selection and reduction of document content. While these methods have demonstrated the ability to condense information, they frequently overlook deeper document-level structures, resulting in summaries that may lack coherence, logical flow, and a clear representation of the original text’s hierarchy as seen in most discussion forums. {Large Language Model (LLM)-based} multi-agent frameworks have been used in summarization for long-form documents through hierarchical processing \cite{chang2023booookscore, guo2024large, zhang2024chain, kim2025nexussum}. However these methods do not balance dynamic viewpoints, especially in noisy, multi-speaker settings \cite{olabisi2022analyzing}.

{Early work on structured textual threads and conversations explored abstractive summarization across a spectrum of multi-turn settings. \citet{zhang2021emailsum} introduced a hierarchical model for summarizing multi-turn email threads, capturing structure at both the email and token levels. \citet{fabbri2021convosumm} then extended this line to multi-party conversations, integrating argument mining to identify argumentative roles and relationships. \citet{fabbri2022answersumm} carried these ideas into Q\&A forums, proposing a pipeline that combines answer selection, content unit extraction, and abstraction to summarize multiple answers to the same question. Building on this, \citet{wu2025unfolding} framed content selection as an iterative self-questioning process that guides summary generation. More recent work, such as \cite{overbay2023mredditsum}, further broadens the paradigm by incorporating multimodal signals from images into thread summarization.
However, these methods do not employ explicit multi-step reasoning over structure or content units to ensure content coverage and coherence.}

In this work, we are concerned with two main research questions. \textbf{(RQ1): How do we address discourse coverage, especially when multiple interleaved topics are present in the threaded documents?} In addition, in multi-party speech, turns often overlap and speakers interrupt each other, making it difficult to rely on a simple linear adjacency model to infer what response was targeted to a comment, leading to our second research question \textbf{(RQ2): How do we ensure that the overlap and interruptions present in complex threaded discussions are considered in order to generate a coherent summary, even without predefined thread order?}

\begin{figure*}[t]
\centering
\resizebox{0.85\textwidth}{!}{
\begin{tikzpicture}[
    font=\small,
    flow/.style={-Latex, line width=3.2pt, color=myBlue},
    flowthin/.style={-Latex, line width=2.0pt, color=myBlue},
    module/.style={
      draw=myDash,
      dashed,
      rounded corners=6mm,
      line width=1.1pt,
      inner sep=6mm,
      minimum width=6.0cm,
      minimum height=5.2cm
    },
    header/.style={
      rounded corners=4mm,
      draw=black!55,
      line width=0.9pt,
      minimum width=5.2cm,
      minimum height=1.15cm,
      align=center
    },
    item/.style={
      draw=black!55,
      line width=0.8pt,
      minimum width=2.8cm,
      minimum height=0.7cm,
      align=center
    },
    card/.style={
      draw=black!55,
      fill=myLightBlue,
      line width=0.8pt,
      minimum width=3.0cm,
      minimum height=0.75cm,
      align=center
    },
    eval/.style={
      draw=black!55,
      fill=myBlue!85,
      rounded corners=4mm,
      line width=0.9pt,
      minimum width=6.2cm,
      minimum height=1.35cm,
      align=center,
      text=white
    },
    docpage/.style={
      draw=black!70,
      fill=white,
      line width=0.8pt,
      minimum width=1.25cm,
      minimum height=1.55cm
    }
]

\node[cloud, cloud puffs=13, cloud puff arc=115, aspect=2.1,
      draw=black!65, line width=1.0pt, fill=myCloud,
      minimum width=7.6cm, minimum height=5.4cm] (cloud) 
      {};

\foreach \name/\x/\y in {D2/-2.0/1.1, D1/0.1/1.6, D3/2.2/0.7,
                         D5/-2.7/-0.1, D4/-0.1/0.2, D8/2.1/-0.4,
                         D6/-2.0/-1.3, D7/0.2/-1.6}{
  \node[draw=black!60, fill=myCloud!75, line width=0.8pt,
        minimum width=1.0cm, minimum height=1.25cm] (doc-\name)
        at ($(cloud.center)+(\x,\y)$) {};

  \draw[black!60, line width=0.6pt]
        ($(doc-\name.north east)+(-0.22cm,0)$) --
        ($(doc-\name.north east)+(0,-0.22cm)$) --
        ($(doc-\name.north east)+(-0.22cm,-0.22cm)$) -- cycle;

  \foreach \yy in {0.25,0.10,-0.05}{
    \draw[black!55, line width=0.45pt]
      ($(doc-\name.center)+(-0.32cm,\yy)$) -- ($(doc-\name.center)+(0.32cm,\yy)$);
  }

  \node[font=\large\bfseries] at ($(doc-\name.center)+(0,-0.38cm)$) {$\name$};
}

\node[module, right=1.2cm of cloud] (m1) {};
\node[header, fill=myPeach] at ($(m1.north)+(0,-1.0cm)$) {\textbf{Aspect Extraction}\\Extract the unique aspects\\in all the documents};

\node[item, fill=myPeach!70] (a1) at ($(m1.center)+(0,0.2cm)$) {Aspect 1};
\node[item, fill=myPeach!70, below=1mm of a1] (a2) {Aspect 2};
\node[font=\Large] at ($(m1.center)+(0,-0.7cm)$) {$\cdots$};
\node[item, fill=myPeach!70] at ($(m1.south)+(0,0.9cm)$) {Aspect $n$};

\node[module, right=1.2cm of m1] (m2) {};
\node[header, fill=myLightBlue] at ($(m2.north)+(0,-1.0cm)$) {\textbf{ACU Generation}\\Generate all atomic content\\units on each aspect};

\node[card] (c1) at ($(m2.center)+(0,0.35cm)$) {ACU 1};
\node[card, below=1mm of c1] (c2) {ACU 2};
\node[font=\Large] at ($(m2.center)+(0,-0.55cm)$) {$\cdots$};
\node[card] at ($(m2.south)+(0,0.9cm)$) {ACU $r$};

\draw[flow] (cloud.east) -- (m1.west);
\draw[flow] (m1.east) -- (m2.west);

\node[module, below=1.0cm of m2] (m3) {};
\node[header, fill=myLightBlue] at ($(m3.north)+(0,-1.0cm)$) {\textbf{Sentence Reordering}\\Reorder ACUs to flow\\coherently};

\node[card] (r1) at ($(m3.center)+(0,0.35cm)$) {ACU 2};
\node[card, below=1mm of r1] (r2) {ACU 1};
\node[font=\Large] at ($(m3.center)+(0,-0.55cm)$) {$\cdots$};
\node[card] at ($(m3.south)+(0,0.9cm)$) {ACU $r$};

\node[module, left=1.2cm of m3] (m4) {};
\node[header, fill=myGreen] at ($(m4.north)+(0,-1.0cm)$) {\textbf{Paragraph Writing}\\Generate paragraph proposals\\from reordered ACUs};

\node[docpage] (p1) at ($(m4.center)+(-1.3cm,-0.35cm)$) {};
\node[docpage] (pn) at ($(m4.center)+(1.3cm,-0.35cm)$) {};
\node[font=\Large] at ($(m4.center)+(0,-0.1cm)$) {$\cdots$};
\node[font=\large\bfseries] at ($(p1.north)+(0,0.25cm)$) {$P_1$};
\node[font=\large\bfseries] at ($(pn.north)+(0,0.25cm)$) {$P_n$};

\draw[black!70, line width=0.6pt]
  ($(p1.north east)+(-0.22cm,0)$) -- ($(p1.north east)+(0,-0.22cm)$)
  -- ($(p1.north east)+(-0.22cm,-0.22cm)$) -- cycle;
\draw[black!70, line width=0.6pt]
  ($(pn.north east)+(-0.22cm,0)$) -- ($(pn.north east)+(0,-0.22cm)$)
  -- ($(pn.north east)+(-0.22cm,-0.22cm)$) -- cycle;

\foreach \yy in {0.30,0.12,-0.06,-0.24}{
  \draw[black!55, line width=0.5pt]
    ($(p1.center)+(-0.38cm,\yy)$) -- ($(p1.center)+(0.38cm,\yy)$);
  \draw[black!55, line width=0.5pt]
    ($(pn.center)+(-0.38cm,\yy)$) -- ($(pn.center)+(0.38cm,\yy)$);
}

\node[eval, anchor=east] (m5) at ($(m4.west|-m4.north)+(-1.2cm,-1.0cm)$) {%
\textbf{Paragraph Evaluation and Selection}\\
Final Summary $=$ $\max(\mathrm{Coh},\mathrm{Cov})\{P_1,\ldots,P_n\}$
};

\draw[flow] (m2.south) -- (m3.north);
\draw[flow] (m3.west |- m5.east) -- (m4.east |- m5.east);
\draw[flow] (m4.west |- m5.east) -- (m5.east);

\node[
  rectangle,
  draw=black!70,
  fill=white,
  minimum width=1.6cm,
  minimum height=2.1cm,
  line width=0.9pt,
  align=center,
  font=\small
] (summarydoc) at ($(m5.south)+(-2.2cm,-1.5cm)$) {};

\draw[black!70, line width=0.7pt]
  ($(summarydoc.center)+(-0.55cm,0.65cm)$) -- ($(summarydoc.center)+(0.25cm,0.65cm)$);
\foreach \yy in {0.45,0.25,0.05,-0.15}{
  \draw[black!55, line width=0.55pt]
    ($(summarydoc.center)+(-0.55cm,\yy)$) -- ($(summarydoc.center)+(0.55cm,\yy)$);
}

\node[font=\small\bfseries] at ($(summarydoc.south)+(0,-0.45cm)$) {Summary};

\end{tikzpicture}%
}
\caption{Our proposed framework for Summarization of Nested Discourse Threads} 
\label{fig:pipeline}
\end{figure*}
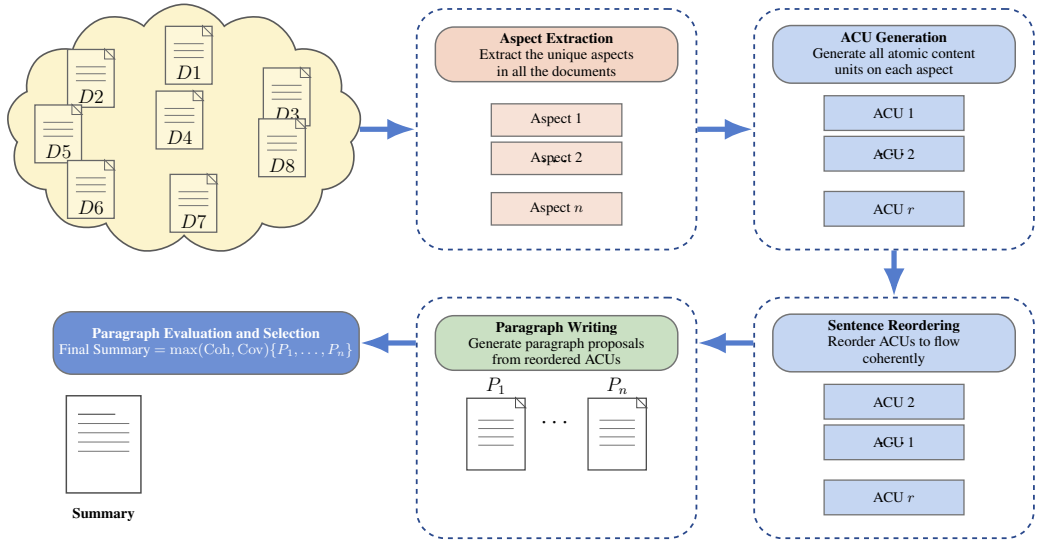

To answer these questions, we make the following contributions:

\begin{itemize}

\item We introduce \textit{ThreadSumm}, a {multi-stage LLM pipeline framework} for threaded multi-document summarization.

\item We employ a thread-aware content-planning layer (Aspects + Atomic Content Units) feeding into a multi-proposal Tree of Thoughts search  to iteratively refine the summary for optimized coherence and coverage.

\item We conduct extensive experiments to show the effectiveness of our method using three LLMs and three datasets across various domains.
\end{itemize}

While our framework integrates established ideas such as Aspect extraction, Atomic Content Units, and LLM-based evaluation, our contribution lies in operationalizing these components using Tree of Thoughts as a single controllable pipeline specifically tailored to nested threaded discourse.

\section{Related Work}

Early neural models incorporated discourse structure into end-to-end architectures: \citet{ishigaki2019dahan} proposed a discourse-aware hierarchical attention network encoding RST dependencies, and \citet{kwon2021neroberta} introduced NeRoBERTa, which injects nested syntactic and discourse trees into a Transformer encoder. In the area of multi-party speech, \citet{zhang2021emailsum} demonstrated that modeling email thread hierarchy improves abstractive summaries, while \citet{fabbri2021convosumm} and \citet{fabbri2022answersumm} proposed pipelines that identify argumentative roles and cluster responses to preserve multiple perspectives in community Q\&A 
discussions. Query-based summarization of discussion threads was explored by \citet{verberne2020query}, who showed that conditioning summaries on user information needs helps surface relevant branches of a discussion. Recent LLM-based frameworks adopt multi-agent or multi-stage processing for long inputs \citep{guo2024modabs,zhang2024chain, kim2025nexussum}. Nevertheless, these approaches either rely on relatively structured interactions or flattened thread structure and optimize for a single-pass summary.

In contrast, \textit{ThreadSumm} explicitly targets unstructured, deeply nested discourse threads by framing summarization as a hierarchical reasoning problem using a thread-aware content-planning layer feeding into a multi-proposal Tree of Thoughts search over paragraph realizations.

\section{ThreadSumm} 
We introduce \textit{ThreadSumm}, a {multi-stage LLM pipeline} framework that summarizes threaded documents using Tree of Thoughts. Figure \ref{fig:pipeline} presents an overview of the process. First, we  decompose content into aspects and atomic units, use Tree of Thoughts reasoning to iteratively reorder and select candidate paragraphs, jointly optimizing coverage and coherence without using predefined thread order. This contrastive design directly addresses the gap between hierarchical summarization of linear documents and the requirements of real-world nested discourse.

Let $D = \{D_1, \ldots, D_m\}$ denote a collection of input documents.  

\begin{itemize}
    \item \textbf{Aspects} $A = \{a_1, \ldots, a_n\}$ denote a collection of $n$ aspects representing the \textit{who, what, where} information.
    \item \textbf{Atomic Content Units (ACUs)} $C = \{c_1, \ldots, c_r\}$, denote a collection of $r$ ACUs, where each unit $c$ is a self-contained semantic statement and $r \geq n$.
\end{itemize}

\begingroup \tabcolsep=4pt\def\arraystretch{1}
\begin{table*}[!t]
\begin{threeparttable}
    \centering
    \small
\begin{tabular}{lccccccc} 
\midrule
Dataset & Domain & \#Instances & Avg Input Length & Avg Summary Length & \#Docs/Example & Ref Summary\\
\midrule
Reddit  & Discussion Forum & 250 & 641 & 127.1 & 7.88 & Y\\
Stack & Community Q\&A & 117 & 1207 & 149.5 & 9.72 & Y\\
Bitcoin \tnote{*} & Bitcoin Forum & 1 & 1250 & 190.7 & 26 & N\\

\bottomrule
\end{tabular}
\begin{tablenotes}
            \item [*] {\textit{included as a case study for thread visualization.}}
        \end{tablenotes}
\caption{Dataset Statistics} \label{tab:dataset}
\end{threeparttable}
\end{table*}
\endgroup

\paragraph{Aspect Extraction}
An important step in summarization is deciding the  \textit{who, what, where} elements that are present in the input document. It encourages coverage across multiple facets of the source document, instead of over-focusing on the most frequent topic. This ultimately reduces information loss on less frequent but important aspects, which is especially valuable in multi-document settings. {We prompt the LLM to extract the aspects in the source document using a few-shot method, the prompt is made available in Appendix \ref{sec:A1}}.

\paragraph{Atomic Content Unit Generation}
After extracting the aspects, we generate Atomic Content Units on each of the aspects. Atomic Content Units are self-contained semantic units that represent a single piece of information within a text that cannot be further broken down without losing its meaning \cite{liu2022revisiting}. This step enables pruning, and it is important because it gives a fine-grained, interpretable basis for judging the key informational elements in the source document. It also ensures broader coverage across aspects ensuring that the salience in the source documents is preserved. {Similar to the preceding step, we apply a few-shot approach to prompt the LLM to generate the ACUs including a detailed list of the ACU generation process. The prompt is made available in Appendix \ref{sec:A2}}.

\paragraph{Sentence Ordering}

To maintain a natural discourse structure, especially given the nested nature of the input documents, we reorder the generated ACUs so that they form a coherent and logical passage - each sentence logically following its predecessor \cite{logeswaran2016sentence, golestani2021using}. This results in a connected narrative rather than a list of disjointed statements, maintaining the logical progression and contextual flow needed for the next step. {We apply a zero-shot method here, prompting the LLM to reorder the ACU list to follow a logical and coherent flow. The prompt is made available in Appendix \ref{sec:A3}}.  
Sentence ordering was done using LLM because of its potential to solve a harder, more global discourse problem than simple heuristics or logic rules can reliably handle. For example, positional/section-based heuristics (e.g., lead sentences first) are not particularly well-suited to threaded discourse representative in our work, where important content appears anywhere in reply chains, not just at the top. Similarly, temporal/chronological rules (e.g., sort by timestamp) could work in principle but would require metadata that is often unavailable in real world settings. Our framework is deliberately metadata agnostic to mimic these constraints, making LLM-based ordering the more robust choice.

\paragraph{Paragraph Writing}
This step takes the reordered sentences from the previous step as input, and generates multiple coherent paragraph proposals using a zero-shot prompt-based method. The model is prompted to start the paragraph with the first sentence in the reordered sentence list, and to end the paragraph with the last sentence in the reordered list. The prompt instructs the model to summarize the key points from the set of ordered sentences into a coherent paragraph. This addresses coherence by turning the set of ordered sentences into a fluent well-structured paragraph that reads as a single cohesive unit rather than a list of sentences. The prompt also explicitly instructs the model to not omit any of the sentences when summarizing the key points into a paragraph. By conditioning on the full set of ordered sentences, the model should integrate and balance all perspectives or aspects mentioned, rather than selectively omitting minority views. The prompt is shown in Appendix \ref{sec:A4}.

We employ the Tree of Thoughts (ToT) approach introduced in \citet{yao2023tree} to iteratively  generate paragraph proposals, selecting the best based on coherence and coverage scores. Coverage measures how well the summary includes the important information in the source document, this is optimized to address \textbf{RQ1}. To address \textbf{RQ2}, we optimize coherence which measures the idea connectedness and logical flow in the summary. {ToT takes a given number of steps $s$ as a parameter, and for each step, it takes multiple proposal generation of sentence reordering and paragraph candidates, and evaluates each candidate against the original source document for its coherence and coverage using an LLM {prompt shown in Appendix \ref{sec:A5}}.  The reordering associated with the highest-scoring paragraph gets carried forward to the next iteration, allowing the model to explore a more diverse set of ''thoughts'' and potentially converge on a higher quality output. This iterative refinement and branching significantly enhances the generation of coherent and comprehensive paragraphs by systematically exploring and evaluating multiple compositional  paths for a more robust search for the best summary.} 

Our method employs multi-step reasoning by decomposing thread summarization into a sequence of structured interdependent reasoning modules tailored specifically to threaded discourse, each step performs a distinct reasoning task with intermediate outputs that are passed on to the next step to generate the final summary

\section{Experimental Setup}

This section describes the implementation details, dataset, baseline methods and evaluation metrics used in our experiments. 

\subsection{Implementation Details}
{We experiment with three LLMs ordered from proprietary to open-source, both larger and smaller models, including GPT-4\footnote{\url{https://developers.openai.com/api/docs/models/gpt-4-turbo}} (\textit{gpt-4-turbo}), Claude\footnote{\url{https://platform.claude.com/docs/en/about-claude/models/overview}} (\textit{claude-3-sonnet})}, and LLaMA 3 (\textit{llama3-70b})\footnote{\url{https://www.llama.com/models/llama-3/}}. 
All detailed prompts and model parameters are shown in Appendix \ref{sec:A}.

\begingroup
\tabcolsep=2.7pt\def\arraystretch{1.2}
\begin{table*}[!t]
\centering
\small
\begin{tabular}{ccccc|ccc|ccc}
\toprule
 \multicolumn{2}{c}{} & \multicolumn{2}{c}\texttt{Reddit}& \multicolumn{3}{c}\texttt{Stack}& \multicolumn{3}{c}\texttt{Bitcoin}\\
 
 \multicolumn{2}{c}{}& {QAGS} & {SummaC}& {ROUGE-1} & {QAGS} & {SummaC}& {ROUGE-1} & {QAGS} & {SummaC}& {ROUGE-1}\\
 
 \midrule
 \multirow{4}{*}{GPT-4} & 
 {Vanilla}& {36.46} & {30.16} & {30.54} & {33.46} & \textbf{35.34} & {32.24}& {71.43} & {48.17}& {5.44} \\
&{arg-graph}& {38.46} & {30.18} & {27.11} & {34.55} & {34.44} & {29.22}& \textbf{85.71} & {57.86}& {4.96} \\
&{CHRONOS}& {41.49} & {30.31} & {28.59} & {33.82} & {35.23} & {31.62}& {71.43} & {46.26}& {6.00} \\
&{mRedditSumm}& {38.23} & \textbf{30.76} & {29.55} & {34.55} & {35.10} & {31.84}& {28.57} & \textbf{61.08}& {5.92} \\
&{ThreadSumm (ours)}& \textbf{50.34} & {30.40} & \textbf{33.30} & \textbf{49.94} & {33.46} & \textbf{35.89}& {57.14} & {40.07}& \textbf{15.28}\\

 \hline 
 \multirow{4}{*}{CLAUDE} & 
{Vanilla}& {38.34} & {29.51} & {30.88} & {40.05} & {36.79} & {31.86}& {28.57} & {45.54}& {4.56}\\
&{arg-graph}& {40.91} & {32.01} & {30.19} & {42.61} & {37.07} & {30.48}& {28.57} & {57.44}& {6.72}\\
&{CHRONOS}& {45.43} & {32.11} & {26.35} & {41.64} & {33.56} & {27.76}& {71.43} & \textbf{67.73}& {4.96} \\
&{mRedditSumm}& {45.66} & {30.64} & {26.82} & {39.93} & {34.87} & {29.22}& {57.14} & {62.28}& {5.20} \\
&{ThreadSumm (ours)}& \textbf{55.66} & \textbf{42.42} & \textbf{34.37} & \textbf{57.75} & \textbf{50.29} & \textbf{39.29} & \textbf{85.71} & {63.72}&\textbf{22.16}\\

 \hline 
 \multirow{4}{*}{LLAMA} & 
{Vanilla}& {33.20} & {28.66} & \textbf{30.53} & {32.84} & {32.46} & {32.69}& {57.14} & {41.33}& {6.08}\\
&{arg-graph}& {35.49} & {30.93} & {24.21} & {36.75} & {38.97} & {27.35}& {28.57} & {49.80}& {6.16}\\
&{CHRONOS}& {43.09} & {28.82} & {27.10} & {31.50} & {36.41} & {28.82}& {42.86} & {31.67}& {5.20} \\
&{mRedditSumm}& {40.34} & {30.34} & {30.37} & {37.36} & {35.00} & {31.96}& {28.57} & {45.52}& {6.24} \\
&{ThreadSumm (ours)}& \textbf{50.00} & \textbf{33.98} & {25.25} & \textbf{51.28} & \textbf{42.44} & \textbf{38.39}& \textbf{71.43} & \textbf{59.36}& \textbf{20.08}\\

\bottomrule
\end{tabular}
\caption{\textbf{\em } Results comparing \textit{ThreadSumm} and other baselines across three models, three datasets, and three evaluation metrics.}
\label{tab:ROUGE}
\end{table*}
\endgroup

\subsection{Dataset}
Table \ref{tab:dataset} presents the overall characteristics of the datasets. In a threaded document setup, one set of documents represents a single thread consisting of one root document or post, plus all its nested replies and sub-replies. We make use of conversational datasets like 
StackOverflow (Stack) \cite{hoogeveen2015cqadupstack}, a community Q\&A forum, and Reddit \cite{zhang2017characterizing}, from CoarseDiscourse. For both datasets, we make use of the preprocessed test set from \citet{fabbri2021convosumm}. We selected the Reddit and StackOverflow datasets from ConvoSumm because they most closely match the discourse characteristics our method is designed to address, namely complex, multi-party threaded discussions with heterogeneous and branching structures. Both Reddit and StackOverflow contain open-ended, community-driven threads in which many participants contribute diverse viewpoints across nested replies, reflecting the core challenges of real-world discussion forums. In contrast, the NYT dataset is centered on responses to a single news article and exhibits a more constrained, less-branching structure, while the Email dataset involves a small number of participants and predominantly linear reply chains. These settings are less representative of the discourse complexity our approach targets.

We also make use of a small amount of data from a bitcoin forum, Bitcoin\footnote{\url{https://bitcointalk.org/index.php}}, which consists of conversation threads, with an average discussion tree depth and breadth of 4 by 6. We include this dataset as a case study as it allows us to perform a more granular analysis given the visual hierarchical structure not included in the other datasets.

\subsection{Baselines}
We conduct experiments using our proposed method, and compare against the following methods which are adapated to use the same underlying LLMs as our method.
\noindent\textbf{Vanilla} - this is a simple prompt-based method using LLMs with the prompt \textit{"You are an expert text summarizer tasked with providing summaries of documents. Please provide a summary of the text. Do not itemize the summary".}
\noindent\textbf{arg-graph} \cite{fabbri2021convosumm} uses  argument mining through graph construction to model the issues and viewpoints, and use a sequence-to-sequence model to generate summaries.
\noindent\textbf{CHRONOS} \cite{wu2025unfolding} uses a retrieval based approach to timeline summarization by iteratively posing questions about the topic and the retrieved documents to generate chronological summaries.
\noindent\textbf{mRedditSumm} \cite{overbay2023mredditsum} uses an approach of comment clustering, cluster summarization, and cluster summary summarization. Although this method was proposed for text and images, we make use of the text-only method.

\subsection{Evaluation Metrics}

We employ the following  metrics for evaluation. \textbf{QAGS} \cite{wang2020asking} uses a question-generating model and a question-answering model to evaluate the quality of generated summaries by detecting factual inconsistencies. \textbf{SummaC} \cite{Laban2022SummaCRN} is an NLI-based method used to assess factual consistency by evaluating whether the generated summary is entailed by the original documents, we report the SummaC-Conv variant. \textbf{ROUGE} \cite{lin2004rouge} calculates the lexical overlap between the machine-generated output and the  human-written reference summaries. To align with coverage evaluation, we report the recall scores of ROUGE-1 (overlapping unigrams) for our experiments.  Since the Bitcoin dataset has no reference summary, we directly use the source document for the overlap as introduced in \citet{bao2023docasref}.

\section{Results and Discussion}

\subsection{Meta Evaluation}
Table \ref{tab:ROUGE} presents the detailed results of our experiments. We observe that \textit{ThreadSumm} consistently outperforms all baselines across all three models (GPT-4, Claude, LLaMA) and all three datasets (Reddit, Stack, Bitcoin) on almost every metric. The performance gains from \textit{ThreadSumm} are substantial, often exceeding baselines by large absolute margins (e.g., +10 to +20 points on QAGS or ROUGE-1), and is the only method that generalizes robustly across domains. Although the performance boost provided by \textit{ThreadSumm} is evident regardless of the underlying LLM used (GPT-4, Claude, or Llama), \textit{ThreadSumm} paired with Claude achieves the best overall performance in the entire table. Even as a smaller and open-source model compared to the others, Llama with \textit{ThreadSumm} remains highly competitive. CHRONOS and mRedditSumm remain the most competitive baselines. {Interestingly, we observe that although the ROUGE scores on the Bitcoin dataset are significantly lower compared to the other datasets, and this can be attributed to the use of the actual source document as the reference document in the calculation, it is worthy of note that \textit{ThreadSumm} still significantly outperforms all other methods, achieving double digit scores, while all the other methods have single digit scores.}
To better align evaluation with our objectives, we complement these standard metrics with targeted analyses of positional bias, aspect retention, and opinion-cluster coverage, and with human ratings focused specifically on coherence and coverage.

\subsection{How does positional bias manifest in nested discourse threads?}

\begin{figure}[!t]
 \centering
 \includegraphics[width=0.3\textwidth]{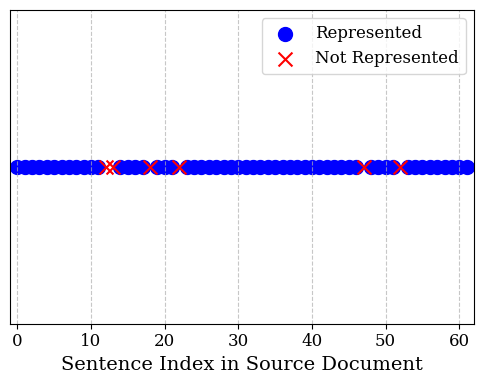}

 \caption {{Represented vs. Not represented sentences using \textit{ThreadSumm} on Bitcoin dataset. The x-axis represents the index of the sentence in the source document.}}
  \label{fig:Position}
\end{figure}

Positional bias is a known challenge in text summarization, this form of bias happens when the model relies on where information appears in the input rather than the content salience, making generated summaries overly sensitive to sentence order \cite{olabisi2024understanding, mayilvaghanan2025spot, ma2025input}. It is essential to control positional bias when summarizing nested discourse threads because posts are often a mix of replies and quotes to either the initial post or replies to the initial post. If a summarizer favors early-position documents, whichever group or perspective that appears first will be overrepresented in the summary thereby creating a distorted view. To evaluate position bias in our method, we compute the proportion of source sentences covered by the summary as a function of their position in the source. {We employ the method in  \cite{sun2025posum} to identify sentences underrepresented or omitted in the generated summary using semantic similarity, softmax normalization, quantile transformation, and a dynamic threshold. A sentence is considered represented if it is greater than or equal to the dynamic threshold.} To better visualize if the summary content is skewed to a certain part of the source document, we plot the position coverage indicating which source sentences are represented in the summary.  Figure \ref{fig:Position} shows the position of represented vs. non-represented sentences in the source document. We see that only 6 of the 62 sentences are not represented, and also that our method does not favor a specific location, with the sentence position not represented spread out. We further analyze the omitted sentences in Appendix \ref{sec:C} and find that they are sentences with no major significance to the key points.

\subsection{Aspect retention in summary}

\begin{figure}[!t]
 \centering
 \includegraphics[width=0.4\textwidth]{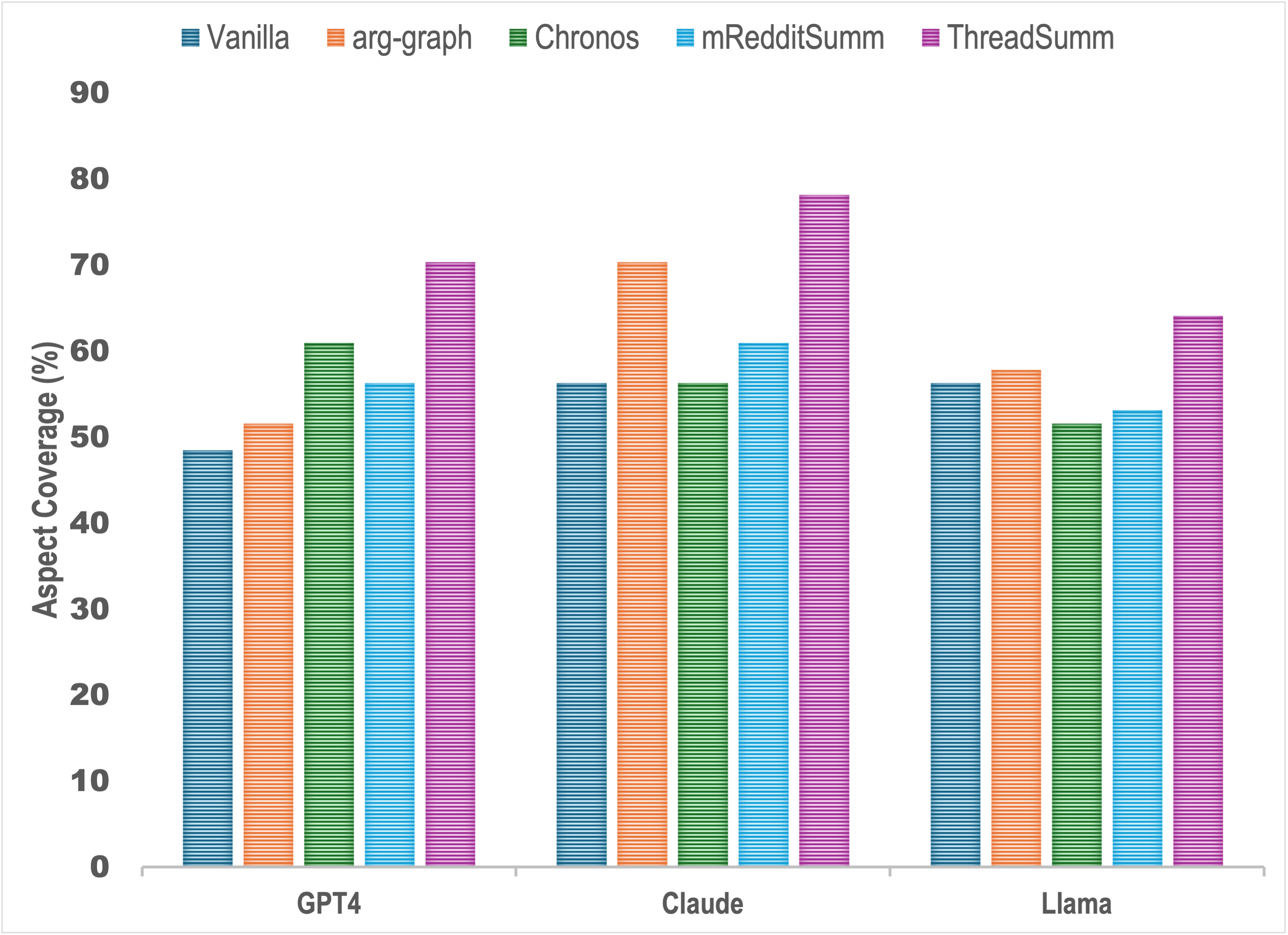}

 \caption {Aspect Coverage of the summary relative to the source document on Bitcoin dataset.}
  \label{fig:Aspect}
\end{figure}

Summarization models should be able to preserve the key elements in the source document without information loss. These key elements (aspects) contain information about the \textit{who, what, where} present in the document. A good summary must cover those aspects rather than just high-frequency surface content. This is closely related to balanced coverage which ensures that summaries include all important aspects instead of over-focusing on just those that occur frequently, as is common in nested discourse threads. To measure this, we compute the aspect overlap between the source document and summaries by extracting the aspects using the prompt \textit{``Extract the unique aspects (places, names, products, features, objects) mentioned in the following document''}. After extracting the aspects from both source document and summary, we count the occurrence of common aspects in both lists and present the final score as a fraction of the common aspects relative to the full aspects in the source document. The results are plotted in Figure \ref{fig:Aspect}, and we observe that our method yields the highest aspect overlap as compared to the baselines across all models. This is as expected as we included aspect extraction as the starting point in our framework to optimize aspect overlap, and this ultimately yields more informative and less biased summaries. In addition, we observe that Claude consistently has the best performance across all methods.

\subsection{Opinion Diversity Analysis}

\begingroup\tabcolsep=3pt\def\arraystretch{1.2}
\begin{table}[!t]
\centering
\small
\begin{tabular}{ccccc}
\midrule

& {} &  \texttt{Reddit} & \texttt{Stack}  & \texttt{Bitcoin}  \\

 \midrule

\multirow{3}{*}{GPT-4} &{Vanilla} &{0.35} &{0.49} &{0.40} \\
 &{arg-graph} &{0.37} &{0.48} &{0.40} \\
 &{CHRONOS} &{0.34} &{0.51} &{0.60} \\
 &{mRedditSumm} &{0.34} &{0.50} &{0.40} \\
 &{ThreadSumm (ours)} &\textbf{0.54} &\textbf{0.58} &\textbf{0.60} \\

 \midrule

\multirow{3}{*}{CLAUDE} &{Vanilla} &{0.40} &{0.51} &{0.40} \\
 &{arg-graph} &{0.39} &{0.51} &{0.60} \\
 &{CHRONOS} &{0.34} &{0.45} &{0.60} \\
 &{mRedditSumm} &{0.34} &{0.43} &{0.40} \\
 &{ThreadSumm (ours)} &\textbf{0.64} &\textbf{0.75} &\textbf{0.80} \\

 \midrule

\multirow{3}{*}{LLAMA} &{Vanilla} &{0.33} &{0.45} &{0.40} \\
 &{arg-graph} &\textbf{0.34} &{0.48} &{0.60} \\
 &{CHRONOS} &{0.30} &{0.42} &{0.20} \\
 &{mRedditSumm} &{0.32} &{0.48} &{0.60} \\
 &{ThreadSumm (ours)} &{0.32} &\textbf{0.62} &\textbf{0.80} \\
\bottomrule
\end{tabular}
\caption{Opinion analysis comparing ThreadSumm and other baselines across the three models and datasets.}
\label{tab:Opinion}
\end{table}
\endgroup 

Coverage across distinct viewpoints is essential in nested threaded documents to ensure that not only the most popular or repeated viewpoints are included in the summary. {Ideally, a good summarizer should be able to capture the opinion of the original poster and the diverging opinion of the responders. To evaluate the performance of our method on opinion diversity coverage, we conduct an analysis to compare the opinion diversity in the generated summary to the source documents.} We achieve this by clustering source sentences into opinion clusters using sentence embeddings and K-means using \textit{all-MiniLM-L6-v2} from the Sentence Transformers family. A cluster is considered covered if at least one summary sentence is semantically aligned with a source sentence in that cluster, beyond a similarity threshold. We set the clusters \textit{k} = 5 to account for a variety of stances/viewpoints, and similarity threshold \textit{t} = 0.6. In this sense, coverage is used as a measure of opinion diversity preservation, ensuring that the summary is touching all the different positions that exist, not just the prominent opinions. The final coverage is computed as the fraction of source clusters represented in the summary. We compare our method against the baselines, and the results in Table \ref{tab:Opinion} show that \textit{ThreadSumm}  
mostly achieves better opinion coverage compared to other methods. {This indicates that our method performs better at preserving opinion balance rather than factual redundancy.}

\subsection{Human Evaluation}

\begingroup\tabcolsep=4pt\def\arraystretch{1.2}
\begin{table}[!t]
\centering
\small
\begin{tabular}{ccccc}
\midrule

& {} &  \texttt{Coverage} & \texttt{Coherence}   \\

 \midrule

 &{Vanilla} &{3.4} &{3.5} \\
 &{arg-graph} &{3.5} &{3.3} \\
 &{CHRONOS} &{3.7} &{3.7} \\
 &{mRedditSumm} &{3.0} &{3.1} \\
 &{ThreadSumm (ours)} &\textbf{4.9} &\textbf{4.6} \\

 \bottomrule
\end{tabular}
\caption{Human Annotation results comparing our method to other baselines.}
\label{tab:Human}
\end{table}
\endgroup

We also collect human evaluation on the three datasets to compare the summaries generated from \textit{ThreadSumm} versus the baselines. Since our research goal is primarily focused on improving discourse coverage and coherence, we use the same two quality dimensions for human evaluation. {We ask the annotators to rate the summaries with respect to the source document on a scale of 1-5 (1 being the worst, and 5, the best).} We provide clear guidelines to the annotators, as shown in Appendix \ref{sec:B}. {We collect evaluation from two annotators for each summary, and compute the inter-annotator agreement on coherence and coverage provided by the annotators using  Krippendorff's alpha.} {The result in Table \ref{tab:Human} shows that our method performed better on both coherence and coverage with medium to high inter-annotator agreement of 0.67 and 0.82.}

\begingroup
\tabcolsep=2.7pt\def\arraystretch{1.2}
\begin{table*}[!t]
\centering
\small
\begin{tabular}{ccccc|cccc}
\toprule
 {} & \multicolumn{4}{c}\texttt{Uncontrolled Length (number of words)}& \multicolumn{4}{c}\texttt{Controlled Length (number of words)}\\
 
{}& {Length} & {QAGS} & {SummaC}& {ROUGE-1} & {Length} & {QAGS} & {SummaC}& {ROUGE-1} \\
 
 \midrule
{Vanilla}& {107} & {37.29} & {30.71} & {31.53} & {76} & {28.12} & {25.19} & {24.01}\\
{arg-graph}& {104} & {39.47} & {31.28} & {33.62} & {71} & {31.29} & {27.03} & {23.72}\\
{CHRONOS}& {103} & {43.22} & {30.85} & {34.91} & {70} & {37.85} & {27.61} & {25.29} \\
{mRedditSumm}& {117} & {44.93} & {34.12} & {36.44} & {78} & {39.37} & \textbf{29.55} & {26.38}\\
{ThreadSumm (ours)}& {169} & \textbf{52.74} & \textbf{34.79} & \textbf{40.18} & {73} & \textbf{45.12} & {26.41} & \textbf{27.93}\\

\bottomrule
\end{tabular}
\caption{\textbf{\em } Summary Length Analysis}
\label{tab:Length}
\end{table*}
\endgroup

\subsection{Summary Length Sensitivity}
We run a controlled experiment with summaries set to length = 5 sentences and compute new scores in Table \ref{tab:Length}. We observe that shortening the summaries lowers performance on all metrics, with ROUGE being the most sensitive to length reduction, followed by QAGS, while SummaC changed only marginally. However, under both settings, controlled length as well as uncontrolled length, \textit{ThreadSumm} mostly outperforms other methods, except on SummaC where mRedditSumm has the highest score. In addition, we conducted a brief analysis comparing our method to mRedditSumm on the Bitcoin dataset under reduced depth and breadth of the input document set, and observed comparable performance between both methods only when both depth and breadth of the document set were limited to 2. This further shows the advantage of our method on more complex, deeply nested hierarchical discourse structures.

\subsection{Qualitative Analysis}

\begingroup\tabcolsep=4pt\def\arraystretch{1.2}
\begin{table*}[!t]
    \centering
    \small
    \begin{tabular}{p{\textwidth}}
        \midrule
   {\textbf{Input} : \texttt{Current car is surely on its way out.  Too much rust and too much shit going wrong with it to warrant all the repairs that it needs.  I found a 1988 Mazda 323 on Craigslist with sub-30k miles on it in pretty much new condition.      What's the most you'd pay for something like this?    </s>I might buy it first...  Late 80's were epic for Japan.  Keep on the look out for an AZ-1 as well ;)   </s>I wouldn't pay \$4900, that's for sure.      </s>Definitely not ;)   </s>People ask for outrageous sums for these low mileage cars that nobody cares about. I'm guessing the same is the case here?..... \textbf{(truncated 20 lines)}}}\\  
   \midrule
   {\textbf{Vanilla Summary} :  \texttt{The user is considering buying a 1988 Mazda 323 with low mileage, found on Craigslist, but thinks the current price is too high. The car has been barely used, with a little old lady driving it only once a week, but this lack of use may have led to neglect of regular maintenance, such as replacing the timing belt. The user and others discuss what a fair price for the car would be, with suggestions ranging from \$2,000 to \$3,500, and some expressing a fondness for 80s cars, including the Mazda 323, for their reliability and low maintenance costs.}}\\

    \midrule
   {\textbf{mReddit Summary} : \texttt{A person is considering purchasing a 1988 Mazda 323 with low mileage, but is hesitant due to the high asking price and potential maintenance costs. They would actually prefer a higher-mileage vehicle with a proven track record of maintenance. The initial asking price was \$7,000, which was later reduced to \$5,000, but most people agree that this is still too high. A more reasonable price for the vehicle would be between \$2,000 to \$3,500. The buyer is waiting to see if the price drops further and is seeking more information before making a decision.}}\\

    \midrule
    {\textbf{ThreadSumm Summary} : \texttt{\textcolor {green} {Japan} produced many epic cars in the late 80s. Mazda 323 is a car model from the late 80s. The AZ-1 is another car model from Japan. Interestingly, a notable example of a well-preserved car from this era is the 1988 Mazda 323, which was found on \textcolor {green}{Craigslist}. The 1988 Mazda 323 has \textcolor {green}{sub-30k miles} on it and is in pretty much new condition. This remarkable condition can be attributed to the fact that the car has been maintained over the years by a \textcolor {green}{little old lady}, and the car has been driven \textcolor {green}{only once a week for years}. Despite its good condition, the car will need a complete service and \textcolor {green}{timing belt replacement}. The Mazda 323 was initially listed for nearly \$7k, which is around the \textcolor {green}{same price as an RX7}. However, the \textcolor {green}{listing price} of the Mazda 323 has been lowered to \$5k. In comparison, in \textcolor {green}{Australia}, a Mazda 323 in good condition would cost around \$2 to \$3 thousand. It's worth noting that some people prefer a well-maintained car with \textcolor {green}{200,000 miles} over a low-mileage car.}}\\

    \bottomrule
    \end{tabular}
    \caption{{Example showing an input document from the Reddit dataset, the summaries generated using the vanilla prompt, mReddit method, and our method. The words highlighted in green are the key points included using our method, but omitted in either the vanilla or mReddit method.}}
    \label{tab:Qualitative}
\end{table*}
\endgroup

{We manually analyze the summaries qualitatively in addition to the automated evaluation. Table \ref{tab:Qualitative} shows a sample source document, the corresponding summary generated using the vanilla prompt,  mReddit method, and the summary generated using our method. We observe that the vanilla method gives a very high-level summary, omitting key points like the current listing price, car mileage, and country of production. This is also observed in the mReddit summary, where there is no mention of the platform where the car was listed, the car mileage or country of production. Our method, on the other hand, covers a lot more key points in the source document (highlighted in green font), yielding longer summaries, while still maintaining a coherent flow of information from the first sentence to the last sentence. Also notable is the fact that slangs such as ``grand'' (thousand), and acronyms such as ``AUS'' (Australia) are not omitted, but properly represented in the summary.}

\subsection{Ablation}

\begingroup\tabcolsep=2pt\def\arraystretch{1.2}
\begin{table}[!t]
\centering
\small
\begin{tabular}{ccccc}
\midrule

& {} &  \texttt{QAGS} & \texttt{SummaC}  & \texttt{ROUGE-1}  \\

 \midrule

 &\texttt{ThreadSumm} &{85.71} &{63.72} &{22.16}  \\

 \midrule

&\texttt{w/o Aspect Extraction} &{57.14} &{50.19} &{21.36}  \\

 &\texttt{w/o Sentence Reordering} &{57.14} &{45.06} &{21.04} \\

 &\texttt{w/o Tree Of Thought} &{57.14} &{44.44} &{6.08} \\
\bottomrule
\end{tabular}
\caption{{Ablation study of the necessity of Aspect Extraction, Sentence Reordering and Tree of Thoughts}}
\label{tab:Ablation1}
\end{table}
\endgroup

\begingroup\tabcolsep=2pt\def\arraystretch{1.2}
\begin{table}[!t]
\centering
\small
\begin{tabular}{ccccc}
\midrule

& {} &  \texttt{QAGS($\Delta$)} & \texttt{SummaC($\Delta$)}  & \texttt{R-1($\Delta$)}  \\

 \midrule

\multirow{3}{*}{GPT4} &\texttt{r = 2, p =2} &{2.38} &{14.29} &{-4.56}  \\

 &\texttt{r = 1, p =1} &{9.79} &{-14.28} &{-3.12} \\
 &\texttt{r = 2, p =1} &{1.62} &{42.86} &{-3.36} \\

 \midrule

\multirow{3}{*}{CLAUDE} &\texttt{r = 2, p =2} &{14.16} &{-14.28} &{2.72} \\

&\texttt{r = 1, p =1} &{13.71} &{14.29} &{2.4} \\
 &\texttt{r = 2, p =1} &{12.49} &{-57.14} &{-2.64} \\

 \midrule

\multirow{3}{*}{LLAMA} &\texttt{r = 2, p =2} &{-11.54} &{0} &{-2.16} \\
 &\texttt{r = 1, p =1} &{-10.73} &{14.28} &{-2.88} \\
 &\texttt{r = 2, p =1} &{3.02} &{14.28} &{-2.72} \\
\bottomrule
\end{tabular}
\caption{{Delta observed in varying reordering proposals (\texttt{r}) and paragraph proposals (\texttt{p}) compared to \texttt{r = 1, and p = 2} used in our experiments.}}
\label{tab:Ablation}
\end{table}
\endgroup 

\paragraph{{Effect of Aspect Extraction, Sentence Reordering, and Tree of Thoughts}}

{To assess the contribution of \textbf{Aspect Extraction}, we remove this component from our pipeline and report results on the Bitcoin dataset with the CLAUDE model (Table~\ref{tab:Ablation1}). While the ROUGE‑1 drop is under 1 point, QAGS and SummaC degrade sharply, indicating that Aspect Extraction is critical for factual consistency and overall summary quality.} {To measure the effect of \textbf{Sentence Reordering}, we compare our framework without this step. We see that removing Sentence Reordering leads to an even larger decline in SummaC and ROUGE‑1 than omitting Aspect Extraction, confirming that reordering sentences before generation is important for producing a coherent, connected narrative.} {Finally, we ablate the \textbf{Tree of Thoughts} step. Although QAGS remains comparable to the settings without Aspect Extraction and Sentence Reordering, both SummaC and ROUGE-1  drop substantially when Tree of Thoughts is removed. This suggests that Tree of Thoughts is the single most impactful component in our pipeline, yielding the largest overall gains.}

\paragraph{Effect of Number of Proposals}To assess the effect of the number of proposals in the ToT step, we analyze how our method performs with varying number of 
proposal parameters. Table \ref{tab:Ablation}  provides an initial sensitivity analysis over the number of reordering and paragraph proposals on the Bitcoin case study. While we observe that different configurations can shift QAGS, SummaC, and ROUGE in non-uniform ways, the gains from significantly larger search are not consistently monotonic across metrics.  These mixed patterns suggest that more aggressive ToT search does not automatically translate into better summaries, and that different evaluation metrics may favor different regions of the ToT hyperparameter space.

\section{Conclusion}
We introduce  \textit{ThreadSumm}, a {multi-stage LLM pipeline} framework consisting of Aspect extraction, ACU generation, Sentence ordering, Paragraph writing, Paragraph evaluation and selection using Tree of Thoughts for nested discourse summarization. Our method optimizes summary coherence and coverage of the source document without needing predefined thread order. We conduct extensive experiments across models and forum datasets, and our results show that our proposed method offers significant improvement compared to the baseline methods, while achieving higher aspect retention and opinion coverage in nested discussions. {Since we focused on optimizing only coherence and coverage in this work, for future work, we will extend the Tree of Thoughts approach to optimize other dimensions of summary quality considering the quality-cost tradeoff. We will also explore how this method can be used in other areas of text summarization aside from discussion forums.} 

\newpage

\section*{Limitations}
{We acknowledge certain limitations that may affect the performance and generalization of our method. Our method does not consider multimodal datasets, as we focused on datasets with textual data specifically English threads only. Also, our Tree of Thoughts search relies on LLM-based coherence and coverage scores, which could introduce potential biases and instability in the evaluation signal. We treat these scores as black box outputs and do not systematically calibrate or validate them against human judgments. Because the same LLMs are used both to generate summaries and to score them, our evaluation loop may inherit model-specific preferences.  Furthermore, our discussions related to the performance of prompting of large language models in generating summaries of nested discourse threads are based solely on the summaries that were developed during this study and the summarization models that were adopted in our experiments. }

\section*{Ethical Considerations}
All the datasets and models we used in this paper are publicly available and accessible. The datasets used in this study are sourced from social forums which are known to contain offensive language; therefore, it is possible for the models to also generate summaries with offensive words.

\section*{Acknowledgments}
We thank the anonymous reviewers as well as the members of PortNLP lab for their valuable feedback. This research was partially supported by the National Science Foundation grant CRII:RI 2246174.


\bibliography{main}

\appendix

\section{Experimental Details} \label{sec:A}
In this section, we provide more experimental details and LLM prompts used. The full algorithm for our method is in Algorithm \ref{alg:Alg1}. We set the temperature at 0 for all LLMs used in this study to ensure that the models generate a deterministic summary. All experiments were run using Nvidia A100 GPU. ThreadSumm introduces additional LLM calls for aspect extraction, ACU generation, sentence ordering, and Tree-of-Thoughts evaluation, which increases computational cost relative to single-pass prompting. On average, summarizing a single thread in our setting requires 5-8 LLM calls depending on the number of steps and proposal, resulting in higher cost of up to 1.5X vs Vanilla method, but yielding ROUGE improvements of up to 65\%, particularly for complex nested discourse structures. For ToT hyperparameter selection, we selected 25 samples from the Reddit dataset and 12 samples from the Stack dataset for the preliminary pilot run, and we picked the configuration with the best mean evaluation performance for our experimental setup: number of steps = 3, number of reordering proposals = 1, and number of paragraph proposal = 2, chosen to balance summary quality and computational cost. We do not claim that our chosen depth and branching factors are globally optimal for all datasets or models.

\subsection{Aspect Extraction} \label{sec:A1}
We use the prompt in Figure \ref{AspectPrompt} across all models to extract all the key aspects in the Source Document.

\begin{tcolorbox}[colback=lightbrown,colframe=black!50!brown,title=Aspect Extraction Prompt,fonttitle=\bfseries]
{You are an expert at understanding documents.}
{Extract the unique aspects (places, names, products, features, objects) mentioned in the following document : \{Input\}}
\end{tcolorbox}
\captionof{figure}{Aspect Prompt}
\label{AspectPrompt}

\subsection{ACU Generation Prompt} \label{sec:A2}
We use the prompt in Figure \ref{ACUPrompt} across all models to generate ACUs on all aspects extracted in the previous step.
Human evaluation was done on a subset of the data to verify the faithfulness of Aspect Extraction and ACU generation, across different prompts.

\begin{tcolorbox}[colback=lightbrown,colframe=black!50!brown,title=ACU Generation Prompt,fonttitle=\bfseries]
SystemPrompt : You are an expert at analyzing documents and breaking them down into atomic content units (ACUs).
An ACU is the smallest self-contained statement of fact, claim, or idea that cannot be further broken down
without losing its meaning.

UserPrompt : Follow this process step by step:

1. Read the document carefully.

2. Read the Aspect list carefully.

3. For each aspect, identify all distinct factual claims, propositions, or ideas.

4. Each ACU should express only one idea.

5. Each ACU should be independent of surrounding text.

6. Each ACU should be written in clear and concise language.

7. Rewrite each idea as a minimal, standalone statement.

8. Do not include reasoning steps or explanations, only the extracted statements.

9. Output the ACUs in a list, where each item is one ACU string.

10. Return only the file as output.
\end{tcolorbox}
\captionof{figure}{ACU Generation Prompt}
\label{ACUPrompt}

\subsection{Sentence Ordering} \label{sec:A3}
We use the prompt in Figure \ref{OrderingPrompt} to reorder the ACUs generated in the previous sentence to form a coherent flow.

\begin{tcolorbox}[colback=lightbrown,colframe=black!50!brown,title=Sentence Ordering Prompt,fonttitle=\bfseries]
{You are an expert at reordering documents for them to follow a logical and coherent flow.}
{Ensure every sentence appears exactly once (none omitted or duplicated). Do not add new content. The reordered output should be a single continuous text, not an itemized list.}
\end{tcolorbox}
\captionof{figure}{Sentence Ordering Prompt}
\label{OrderingPrompt}

\subsection{Paragraph Writing} \label{sec:A4}
We use the prompt in Figure \ref{ParagraphPrompt} across all models to summarize all the key points in the reordered sentences into a succinct paragraph. 

\begin{tcolorbox}[colback=lightbrown,colframe=black!50!brown,title=Paragraph Writing Prompt,fonttitle=\bfseries]
{SystemPrompt: You are an expert at reading documents and summarizing the key points into one coherent paragraph.}

{UserPrompt: Given the following sentences, write a single coherent paragraph (not in bullet points or itemized.
The first sentence of the paragraph must be the first sentence of the input document, and the last sentence of the paragraph must be the last sentence of the input document. Do not omit or duplicate any sentences. Use natural transitions and make it flow well.}
\end{tcolorbox}
\captionof{figure}{Paragraph Writing Prompt}
\label{ParagraphPrompt}

\subsection{Iterative Evaluation} \label{sec:A5}
We use the prompt in Figure \ref{EvaluationPrompt} to evaluate the paragraph candidates against the original text for its coherence and coverage. 

\begin{tcolorbox}[colback=lightbrown,colframe=black!50!brown,title=Evaluation Prompt,fonttitle=\bfseries]
{SystemPrompt: You are an expert at evaluating text for coherence and coverage.}

{UserPrompt: "Given the original text (representing a set of core ideas) and a rewritten paragraph, score the paragraph for:

1. Coherence (logical flow, readability, and overall structure, ranging from 0.0 to 1.0)
2. Coverage (how completely it includes all key ideas/sentences from the original text, ranging from 0.0 to 1.0)
Return the two scores as two numbers separated by a space (e.g., '0.9 1.0'). If the paragraph contains significantly fewer or more sentences than the original text, or if it changes the core meaning, score coverage lower. Do not provide any other text besides the two scores.}
\end{tcolorbox}
\captionof{figure}{Evaluation Prompt}
\label{EvaluationPrompt}

\begingroup\tabcolsep=1pt\def\arraystretch{0.4}
\begin{algorithm}[H]
\small
\caption{ThreadSumm}\label{alg:Alg1}

\begin{algorithmic}[1]

    \STATE Input: Documents $D$
    \STATE Output: Summary $S$
    \STATE $p \gets \text{ConstructAspectPrompt}(D)$
    \STATE $\text{raw} \gets \text{LLM}_{\text{FS}}(p, E)$
    \STATE $A \gets \text{NormalizeAspects}(\text{raw})$

    \STATE $C \gets \emptyset$
    \FOR{each $a \in A$}
        \STATE $p \gets \text{ConstructACUPrompt}(D, a)$
        \STATE $\text{raw} \gets \text{LLM}_{\text{FS}}(p, E)$
        \STATE $C \gets C \cup \text{ExtractACUs}(\text{raw})$
    \ENDFOR
    \STATE $C \gets \text{Deduplicate}(C)$

    \STATE $\text{scores} \leftarrow \text{ComputeCoherenceMatrix}(C)$
    \STATE $\text{start} \leftarrow \text{SelectMostCoherent}(C, scores)$
    \STATE $\text{ord} \leftarrow \text{start}$
    \WHILE {unplaced ACUs remain}
    
       \STATE $\text{next} \leftarrow \text{SelectBestFollower}(\text{ord.last}, C, scores)$
       \STATE $\text{Append(ord, next)}$
    \ENDWHILE

    \STATE $P \leftarrow \emptyset$
    \FOR {i = 1 to K }
        \STATE $p \leftarrow GenerateParagraph($
                \STATE $\text{start} = \text{ord[1]}$,
                \STATE $\text{end}   = \text{ord[last])}$
        \STATE $P \leftarrow P \cup {p}$
    \ENDFOR
    
    \FOR {each paragraph p in P}
        \STATE $coh \leftarrow \text{ComputeCoherence}(p)$
        \STATE $cov \leftarrow \text{ComputeCoverage}(p)$
        \STATE $\text{score}[p] \leftarrow \text{CombineScores}(coh, cov)$
    \ENDFOR
    \STATE $S \leftarrow ArgMax(\text{score})$
    \STATE return $S$

\end{algorithmic}  
\end{algorithm}
\endgroup

\section{Annotation Guidelines} \label{sec:B}

We provide the two English-speaking student annotators, familiar with community forums with clear guidelines on how to conduct evaluation of the samples, shown in Figure \ref{HumanEvaluation}.

\begin{tcolorbox}[colback=lightbrown,colframe=black!50!brown,title=Annotation Guidelines,fonttitle=\bfseries]

\textit{Given the following summaries, and the document, read the content and the summaries properly, and rate them on the following dimensions.}
\vspace{1em}

Coherence: On a scale of 1-5, is the summary well-structured, grammatical and easy to follow?

Coverage: On a scale of 1-5, how well does the summary cover all the points in the content provided?

\end{tcolorbox}
\captionof{figure}{Annotation Guidelines}
\label{HumanEvaluation}

\section{LLM-as-a-judge evaluation} \label{sec:C}

We carried out additional analyses using LLM-as-a-judge with a different model “gpt-oss-20b”. Our results in Table \ref{tab:LLMjudge} show that even under this new metric, \textit{ThreadSumm} outperforms the other methods on Coherence and Coverage, and mRedditSumm remains the most competitive baseline.

\begingroup
\tabcolsep=2.7pt\def\arraystretch{1.2}
\begin{table*}[!t]
\centering
\small
\begin{tabular}{ccccc|ccc|ccc}
\toprule
 \multicolumn{2}{c}{} & \multicolumn{2}{c}\texttt{Reddit}& \multicolumn{3}{c}\texttt{Stack}& \multicolumn{3}{c}\texttt{Bitcoin}\\
 
 \multicolumn{2}{c}{}& {QAGS} & {SummaC}& {ROUGE-1} & {QAGS} & {SummaC}& {ROUGE-1} & {QAGS} & {SummaC}& {ROUGE-1}\\
 
 \midrule
 \multirow{4}{*}{Coherence} & 
 {Vanilla}& {8.7} & {8.2} & {8.6} & {8.6} & {8.9} & {8.6} & {8} & {8} & {8}\\
&{arg-graph}& {8.2} & {7.8} & {8.4} & {7.6} & {8.7} & {6.7} & {6} & {8} & {7}\\
&{CHRONOS}& {8.6} & {8.4} & {8.2} & {8.5} & {7.2} & {8.6} & {8} & {9} & {8} \\
&{mRedditSumm}& {8.8} & {8} & {8.6} & {8.2} & {8.7} & {8.3} & {9} & {8} & {8} \\
&{ThreadSumm (ours)}& \textbf{9.1} & \textbf{9.3} & \textbf{8.9} & \textbf{9.1} & \textbf{9.2} & \textbf{8.9} & \textbf{10} & \textbf{10} & \textbf{9}\\

 \hline 
 \multirow{4}{*}{Coverage} & 
{Vanilla}& {7.8} & {7.8} & {7.4} & {7.6} & {8.5} & {7.8} & {7} & {8} & {6}\\
&{arg-graph}& {7.6} & {7.8} & {7.2} & {8.4} & {7.9} & {7.4} & {7} & {7} & {7}\\
&{CHRONOS}& {8} & {7.6} & {7.8} & {8} & {8.8} & {8} & {7} & {8} & {7} \\
&{mRedditSumm}& {8.2} & {8} & {8.2} & {7.9} & {8.9} & {8.5} & {8} & {8} & {8} \\
&{ThreadSumm (ours)}& \textbf{8.8} & \textbf{8.4} & \textbf{8.2} & \textbf{8.6} & \textbf{9} & \textbf{8.5} & \textbf{9} & \textbf{9} & \textbf{8}\\

\bottomrule
\end{tabular}
\caption{\textbf{\em } LLM-as-a-judge results comparing \textit{ThreadSumm} and other baselines across three models, three datasets, and three evaluation metrics.}
\label{tab:LLMjudge}
\end{table*}
\endgroup

\section{Enhanced Vanilla Prompt} \label{sec:D}

The vanilla baseline we used in the main experiment was kept simple to get a fair, reproducible baseline, and approximate how an off-the-shelf user would call the model. However, we run additional experiments with an enhanced vanilla prompt - \textit{"You are an expert text summarizer tasked with providing summaries of nested multi-turn conversations, with multiple replies and quotes. Read through the documents, to understand the logical flow of the conversation. Please provide a summary of the text, ensuring coherence and coverage of all salient points. Do not itemize the summary. Return nothing else."}  We report the results in Table \ref{tab:enhancedVanilla}, and note that while Enhanced Vanilla improves over the simple Vanilla method, our method still achieves superior performance relative to the enhanced baseline.

\begingroup
\tabcolsep=2.7pt\def\arraystretch{1.2}
\begin{table*}[!t]
\centering
\small
\begin{tabular}{ccccc|ccc|ccc}
\toprule
 \multicolumn{2}{c}{} & \multicolumn{2}{c}\texttt{Reddit}& \multicolumn{3}{c}\texttt{Stack}& \multicolumn{3}{c}\texttt{Bitcoin}\\
 
 \multicolumn{2}{c}{}& {QAGS} & {SummaC}& {ROUGE-1} & {QAGS} & {SummaC}& {ROUGE-1} & {QAGS} & {SummaC}& {ROUGE-1}\\
 
 \midrule
 \multirow{6}{*}{GPT-4} & 
 {EnhancedVanilla}& {38.54} & {30.26} & {31.09} & {37.92} & \textbf{38.34} & {33.91} & {71.43} & {49.20} & {6.29}\\
 &{Vanilla}& {36.46} & {30.16} & {30.54} & {33.46} & {35.34} & {32.24}& {71.43} & {48.17}& {5.44} \\
&{arg-graph}& {38.46} & {30.18} & {27.11} & {34.55} & {34.44} & {29.22}& \textbf{85.71} & {57.86}& {4.96} \\
&{CHRONOS}& {41.49} & {30.31} & {28.59} & {33.82} & {35.23} & {31.62}& {71.43} & {46.26}& {6.00} \\
&{mRedditSumm}& {38.23} & \textbf{30.76} & {29.55} & {34.55} & {35.10} & {31.84}& {28.57} & \textbf{61.08}& {5.92} \\
&{ThreadSumm (ours)}& \textbf{50.34} & {30.40} & \textbf{33.30} & \textbf{49.94} & {33.46} & \textbf{35.89}& {57.14} & {40.07}& \textbf{15.28}\\

 \hline 
 \multirow{6}{*}{CLAUDE} & 
{EnhancedVanilla}& {40.44} & {31.76} & {32.93} & {44.65} & {39.13} & {34.42} & {57.14} & {48.39} & {5.61}\\
 &{Vanilla}& {38.34} & {29.51} & {30.88} & {40.05} & {36.79} & {31.86}& {28.57} & {45.54}& {4.56}\\
&{arg-graph}& {40.91} & {32.01} & {30.19} & {42.61} & {37.07} & {30.48}& {28.57} & {57.44}& {6.72}\\
&{CHRONOS}& {45.43} & {32.11} & {26.35} & {41.64} & {33.56} & {27.76}& {71.43} & \textbf{67.73}& {4.96} \\
&{mRedditSumm}& {45.66} & {30.64} & {26.82} & {39.93} & {34.87} & {29.22}& {57.14} & {62.28}& {5.20} \\
&{ThreadSumm (ours)}& \textbf{55.66} & \textbf{42.42} & \textbf{34.37} & \textbf{57.75} & \textbf{50.29} & \textbf{39.29} & \textbf{85.71} & {63.72}&\textbf{22.16}\\

 \hline 
 \multirow{6}{*}{LLAMA} & 
{EnhancedVanilla}& {35.87} & {30.59} & \textbf{32.35} & {34.77} & {36.42} & {33.53} & {57.14} & {44.28} & {6.26}\\
 &{Vanilla}& {33.20} & {28.66} & {30.53} & {32.84} & {32.46} & {32.69}& {57.14} & {41.33}& {6.08}\\
&{arg-graph}& {35.49} & {30.93} & {24.21} & {36.75} & {38.97} & {27.35}& {28.57} & {49.80}& {6.16}\\
&{CHRONOS}& {43.09} & {28.82} & {27.10} & {31.50} & {36.41} & {28.82}& {42.86} & {31.67}& {5.20} \\
&{mRedditSumm}& {40.34} & {30.34} & {30.37} & {37.36} & {35.00} & {31.96}& {28.57} & {45.52}& {6.24} \\
&{ThreadSumm (ours)}& \textbf{50.00} & \textbf{33.98} & {25.25} & \textbf{51.28} & \textbf{42.44} & \textbf{38.39}& \textbf{71.43} & \textbf{59.36}& \textbf{20.08}\\

\bottomrule
\end{tabular}
\caption{\textbf{\em } Results comparing \textit{ThreadSumm} and other baselines across three models, three datasets, and three evaluation metrics, including enhanced vanilla prompt.}
\label{tab:enhancedVanilla}
\end{table*}
\endgroup

\section{Analysis of sentences not represented in summary} \label{sec:E}

In this section, we analyze the sentences in the Bitcoin source document that are not represented in the summary. A sentence is considered represented if it is greater than or equal to the dynamic threshold which is calculated based on the statistical properties of each conversation-summary pair. Table \ref{tab:NotRepresented} shows the \textit{ThreadSumm} summary, as well as the index and corresponding text not represented in the summary. It is obvious that the underrepresented text does not improve or diminish the summary content in any way, indicating that our method does not show position bias.

\begingroup\tabcolsep=3pt\def\arraystretch{0.6}
\begin{table}[!t]
\centering
\small
\begin{tabular}{p{0.48\textwidth}}
\midrule
{\textbf{ThreadSumm Summary}} \\
\midrule
{The discussion of anonymous payment methods begins with the consideration of cash, which is generally considered the most anonymous way to pay. Cash is considered the best anonymous payment method, but it has its own set of limitations. However, paying with cash can be more anonymous than paying with Bitcoin, but in some cases, Bitcoin can be more anonymous than cash. Large volumes of cash cannot be stored or sent easily, and banks may ask questions or refuse to release large amounts of cash. Using cash can leave fingerprints unless gloves are worn, and the CIA may investigate cash transactions for forensic analysis, although serial numbers on cash do not serve the purpose of tracing ownership. In contrast to cash, Bitcoin transactions have their own set of characteristics. The blockchain is a permanent record of all Bitcoin transactions. Bitcoin transactions can be traced, but ownership cannot be proven. However, Bitcoins can be mixed and re-mixed to prevent the NSA from identifying a Bitcoin holder, and the NSA can see who has a significant amount of Bitcoins. Paying with mixed Bitcoins can be more anonymous than paying with cash. Using a VPN can make Bitcoin transactions more anonymous. Bitcoin transactions are more anonymous online than in person. The use of Bitcoin for various transactions is also noteworthy. Bitcoins can be used for all transactions, regardless of the amount. Using Bitcoin for small transactions contributes to the freedom of the currency. The blockchain can be used for small transactions without 'polluting' it. Face to face transactions of physical goods can only be traced by the other person seeing and remembering the buyer's face. Phone and email records can be traced, but Trac phones and certain emails cannot be traced. Computers always leave digital traces. Ultimately, the most anonymous transaction is the one that utilizes complete online anonymity measures. The only totally anonymous transaction is for digital goods bought online using Bitcoin with complete online anonymity measures. Buying Bitcoin anonymously is possible but often requires more effort than buying cash. Some exchanges, like btc-e, do not require identification to buy Bitcoin. Cash is generally considered the most anonymous way to pay, but Bitcoin can offer a high level of anonymity when used correctly. }\\
\midrule
{\textbf{Posts and Index from Source Document not represented in the summary}} \\
\midrule 
{\#12: ;D. Of course it is.} \\
{\#13: But unless it's lying around you have to pull it out of your account and if it's a large amount that's pretty tricky ...} \\
{\#18: It just depends!.} \\
{\#22: Whoever you give it to is likely to spend it fast, so most of the time you don't have to worry.} \\
{\#47: Absolutely it is.} \\
{\#52: I'm always surprised when I read that some people want to use BTC to buy a soda.} \\

\bottomrule
\end{tabular}
\caption{Analysis of sentences not represented}\label{tab:NotRepresented}
\end{table}
\endgroup

\section{Aspect Coverage on Reddit and Stack Datasets}

\begin{figure}[!t]
 \centering
 \includegraphics[width=0.4\textwidth]{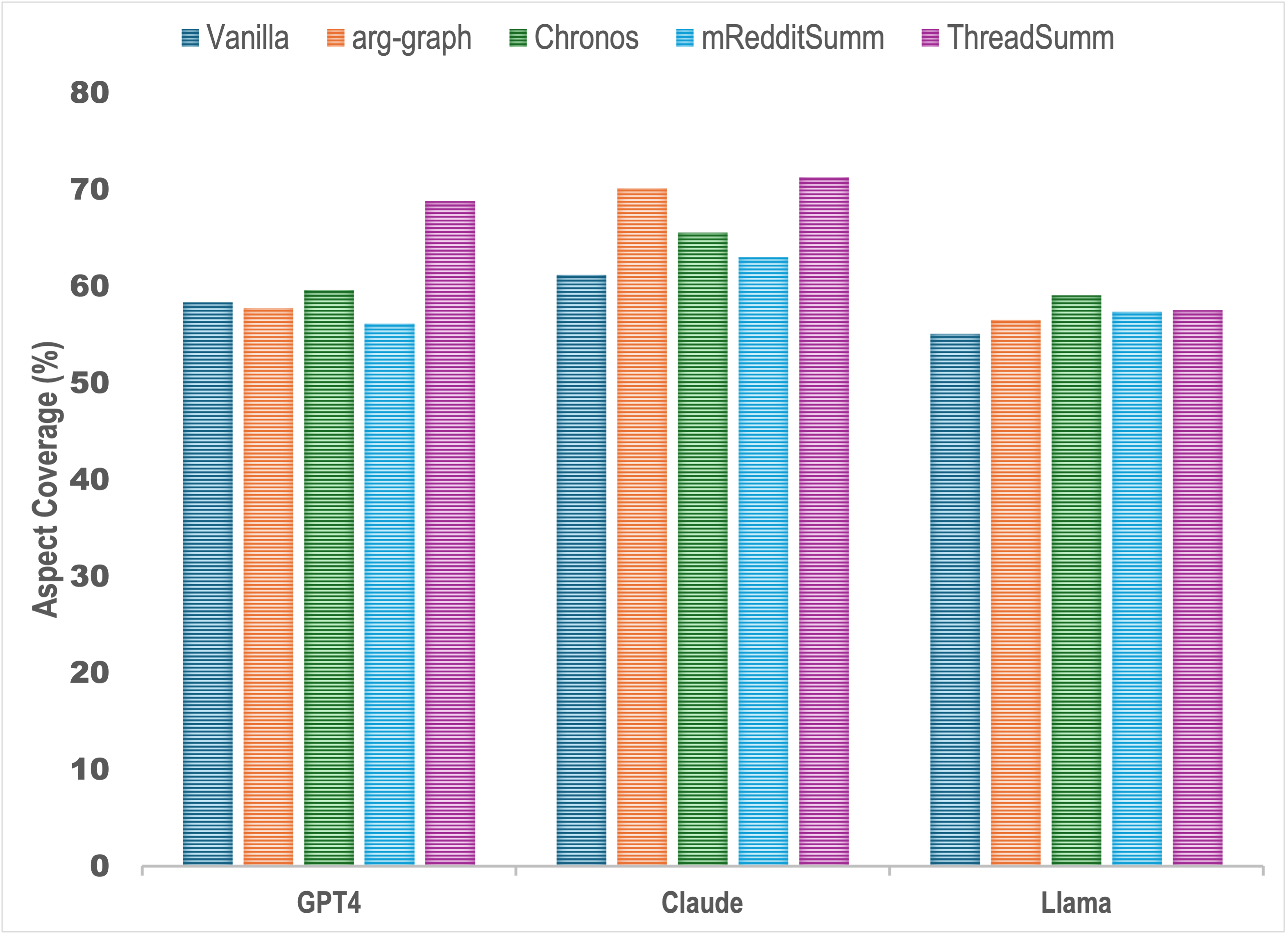}

 \caption {Aspect Coverage of the summary relative to the source document on Reddit dataset.}
  \label{fig:RedditAspect}
\end{figure}

\begin{figure}[!t]
 \centering
 \includegraphics[width=0.4\textwidth]{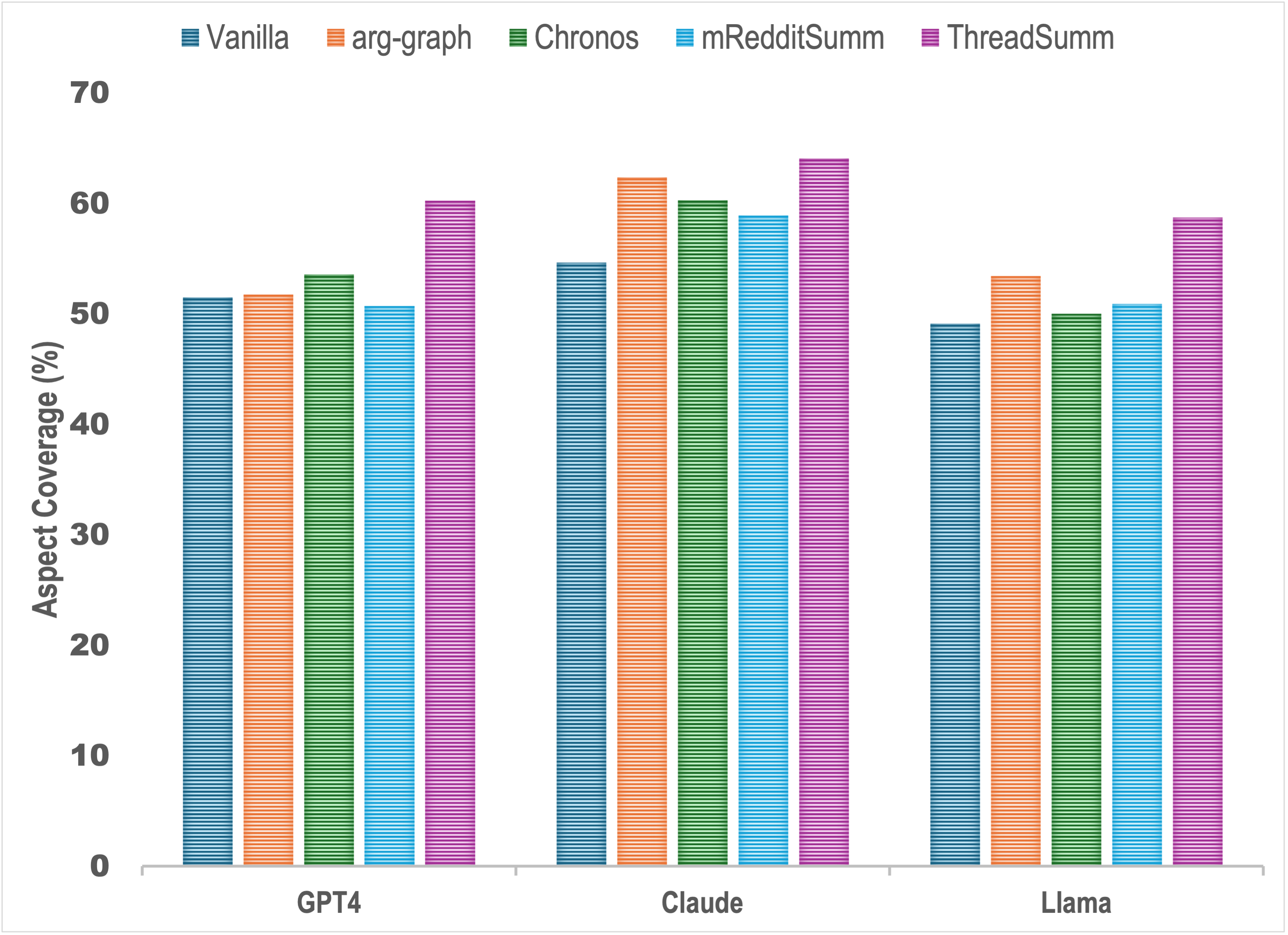}

 \caption {Aspect Coverage of the summary relative to the source document on Stack dataset.}
  \label{fig:StackAspect}
\end{figure}

We plot the Aspect Coverage on Reddit and Stack datasets in Figure \ref{fig:RedditAspect}, and Figure \ref{fig:StackAspect}, and we see that our method mostly outperforms the baseline methods. Claude also has the best performance on both datasets as observed in the Bitcoin datset.

\section{Ablation on Reddit and Stack}
\label{sec:F}

We provide further results in Table \ref{tab:Ablation1Reddit} and Table \ref{tab:Ablation1Stack}, which match the same findings observed on the Bitcoin dataset that Tree of Thoughts is the single most impactful component in our pipeline, yielding the largest overall gains, and in Table \ref{tab:Ablation2Reddit} and Table \ref{tab:Ablation2Stack} showing that different evaluation metrics may favor different regions of the ToT hyperparameter space.

\begingroup\tabcolsep=2pt\def\arraystretch{1.2}
\begin{table}[!t]
\centering
\small
\begin{tabular}{ccccc}
\midrule

& {} &  \texttt{QAGS} & \texttt{SummaC}  & \texttt{ROUGE-1}  \\

 \midrule

 &\texttt{ThreadSumm} &{55.66} &{42.42} &{34.37}  \\

 \midrule

&\texttt{w/o Aspect Extraction} &{48.19} &{31.06} &{30.12}  \\

 &\texttt{w/o Sentence Reordering} &{48.02} &{30.24} &{27.46} \\

 &\texttt{w/o Tree Of Thought} &{39.53} &{28.36} &{25.15} \\
\bottomrule
\end{tabular}
\caption{{Ablation study of the necessity of Aspect Extraction, Sentence Reordering and Tree of Thoughts on Reddit dataset}}
\label{tab:Ablation1Reddit}
\end{table}
\endgroup

\begingroup\tabcolsep=2pt\def\arraystretch{1.2}
\begin{table}[!t]
\centering
\small
\begin{tabular}{ccccc}
\midrule

& {} &  \texttt{QAGS} & \texttt{SummaC}  & \texttt{ROUGE-1}  \\

 \midrule

 &\texttt{ThreadSumm} &{57.75} &{50.29} &{39.29}  \\

 \midrule

&\texttt{w/o Aspect Extraction} &{41.19} &{40.81} &{35.46}  \\

 &\texttt{w/o Sentence Reordering} &{40.63} &{37.22} &{33.75} \\

 &\texttt{w/o Tree Of Thought} &{38.08} &{33.14} &{30.21} \\
\bottomrule
\end{tabular}
\caption{{Ablation study of the necessity of Aspect Extraction, Sentence Reordering and Tree of Thoughts on Stack dataset}}
\label{tab:Ablation1Stack}
\end{table}
\endgroup

\begingroup\tabcolsep=2pt\def\arraystretch{1.2}
\begin{table}[!t]
\centering
\small
\begin{tabular}{ccccc}
\midrule

& {} &  \texttt{QAGS($\Delta$)} & \texttt{SummaC($\Delta$)}  & \texttt{R-1($\Delta$)}  \\

 \midrule

\multirow{3}{*}{GPT4} &\texttt{r = 2, p =2} &{12.67} &{-8.42} &{-3.87}  \\

 &\texttt{r = 1, p =1} &{14.10} &{-10.65} &{-5.18} \\
 &\texttt{r = 2, p =1} &{10.22} &{11.29} &{-3.77} \\

 \midrule

\multirow{3}{*}{CLAUDE} &\texttt{r = 2, p =2} &{11.61} &{-8.49} &{2.91} \\

&\texttt{r = 1, p =1} &{12.39} &{9.34} &{-2.53} \\
 &\texttt{r = 2, p =1} &{10.06} &{-10.40} &{-2.74} \\

 \midrule

\multirow{3}{*}{LLAMA} &\texttt{r = 2, p =2} &{13.72} &{-11.26} &{3.98}  \\
 &\texttt{r = 1, p =1} &{10.12} &{8.31} &{-2.05} \\
 &\texttt{r = 2, p =1} &{10.88} &{9.42} &{2.28} \\
\bottomrule
\end{tabular}
\caption{{Delta observed in varying reordering proposals (\texttt{r}) and paragraph proposals (\texttt{p}) compared to \texttt{r = 1, and p = 2} used in our experiments on Reddit dataset.}}
\label{tab:Ablation2Reddit}
\end{table}
\endgroup

\begingroup\tabcolsep=2pt\def\arraystretch{1.2}
\begin{table}[!t]
\centering
\small
\begin{tabular}{ccccc}
\midrule

& {} &  \texttt{QAGS($\Delta$)} & \texttt{SummaC($\Delta$)}  & \texttt{R-1($\Delta$)}  \\

 \midrule

\multirow{3}{*}{GPT4} &\texttt{r = 2, p =2} &{11.42} &{9.26} &{-2.91}  \\

 &\texttt{r = 1, p =1} &{8.94} &{7.12} &{-4.06} \\
 &\texttt{r = 2, p =1} &{-10.32} &{-10.74} &{-4.17} \\

 \midrule

\multirow{3}{*}{CLAUDE} &\texttt{r = 2, p =2} &{8.63} &{10.29} &{-3.96} \\

&\texttt{r = 1, p =1} &{8.42} &{-8.16} &{2.02} \\
 &\texttt{r = 2, p =1} &{-9.37} &{10.67} &{-2.81} \\

 \midrule

\multirow{3}{*}{LLAMA} &\texttt{r = 2, p =2} &{9.54} &{7.42} &{-2.21}  \\
 &\texttt{r = 1, p =1} &{-5.62} &{10.19} &{-2.64} \\
 &\texttt{r = 2, p =1} &{-8.71} &{9.23} &{-2.11} \\
\bottomrule
\end{tabular}
\caption{{Delta observed in varying reordering proposals (\texttt{r}) and paragraph proposals (\texttt{p}) compared to \texttt{r = 1, and p = 2} used in our experiments on Stack dataset.}}
\label{tab:Ablation2Stack}
\end{table}
\endgroup

\section{Disclosure of use of AI tools}
Portions of this manuscript were prepared with the
assistance of AI tools for editing, and research on related work (OpenAI ChatGPT, Perplexity). These tools were used to improve sentence clarity, while all ideas, analysis, and conclusions are the authors’ own.
\label{sec:appendix}

\end{document}